\documentclass[letterpaper,twocolumn,fleqn]{article} 

\usepackage{ist}
\usepackage{amsmath,amssymb}
\usepackage{graphicx}
\usepackage{booktabs}
\usepackage{multirow}
\usepackage{xcolor}
\usepackage{hyperref}

\title{Motion-Adaptive Temporal Attention for Lightweight Video Generation with Stable Diffusion}

\author{Rui Hong; George Mason University; Fairfax, VA, USA \\
Shuxue Quan; Independent Researcher}

\date{}

\begin{document} 

\maketitle 

\thispagestyle{empty}


\begin{abstract}
We present a motion-adaptive temporal attention mechanism for parameter-efficient video generation built upon frozen Stable Diffusion models. Rather than treating all video content uniformly, our method dynamically adjusts temporal attention receptive fields based on estimated motion content: high-motion sequences attend locally across frames to preserve rapidly changing details, while low-motion sequences attend globally to enforce scene consistency. We inject lightweight temporal attention modules into all UNet transformer blocks via a cascaded strategy---global attention in down-sampling and middle blocks for semantic stabilization, motion-adaptive attention in up-sampling blocks for fine-grained refinement. Combined with temporally correlated noise initialization and motion-aware gating, the system adds only 25.8M trainable parameters (2.9\% of the base UNet) while achieving competitive results on WebVid validation when trained on 100K videos. We demonstrate that the standard denoising objective alone provides sufficient implicit temporal regularization, outperforming approaches that add explicit temporal consistency losses. Our ablation studies reveal a clear trade-off between noise correlation and motion amplitude, providing a practical inference-time control for diverse generation behaviors.
\end{abstract}

\section{Introduction}
\label{sec:intro}

Text-to-video generation has emerged as a frontier challenge in generative modeling. While text-to-image diffusion models such as Stable Diffusion~\cite{rombach2022ldm} have achieved remarkable quality, extending them to temporally coherent video remains difficult. The core challenge is twofold: maintaining visual consistency across frames while producing meaningful motion that aligns with the text prompt.

Existing approaches to video diffusion fall broadly into two categories. Full video diffusion models~\cite{ho2022videodm,singer2022makeavideo,blattmann2023videoldm} train end-to-end on large-scale video datasets, achieving strong results but at enormous computational cost. Parameter-efficient methods such as AnimateDiff~\cite{guo2023animatediff} instead inject trainable temporal layers into frozen image diffusion models, learning motion patterns from video data while preserving the spatial generation quality of the base model. However, these methods treat all video content uniformly, applying identical temporal attention regardless of motion characteristics. This one-size-fits-all strategy is suboptimal: a slow panning landscape and a fast-moving animal require fundamentally different temporal modeling.

We propose a \textbf{Motion-Adaptive Temporal UNet} that addresses this limitation through three key innovations. First, we introduce \textbf{motion-adaptive attention bias}, where the temporal attention mechanism dynamically adjusts its receptive field based on estimated motion content---high-motion sequences attend to nearby frames to preserve detail, while low-motion sequences attend globally to enforce consistency. Second, we employ \textbf{motion-aware gating} that modulates the strength of temporal contributions at each network layer according to motion intensity. Third, we adopt a \textbf{cascaded injection strategy} in which down-sampling and middle blocks apply global temporal stabilization, while up-sampling blocks apply motion-adaptive local refinement.

Our temporal modules are injected into a frozen Stable Diffusion UNet, adding only 25.8M trainable parameters (2.9\% of the base model). Combined with temporally correlated noise initialization, the approach produces temporally coherent videos while adapting to diverse motion patterns. Training is conducted on a 100K subset of WebVid~\cite{bain2021webvid}, demonstrating strong parameter efficiency compared to methods such as AnimateDiff, whose motion modules require 417M parameters. Code is available at \url{https://github.com/hongrui16/motion-adaptive-video}.


\section{Related Work}
\label{sec:related}

\subsection{Latent Diffusion Models}

Latent Diffusion Models (LDMs)~\cite{rombach2022ldm} perform the diffusion process in a compressed latent space learned by a variational autoencoder (VAE), greatly reducing computational cost compared to pixel-space diffusion. Stable Diffusion, an open-source LDM, uses a UNet architecture with spatial self-attention and text-conditioned cross-attention to generate high-quality images from text prompts. Our method builds upon this frozen image generation backbone.

\subsection{Video Diffusion Models}

Early video diffusion models~\cite{ho2022videodm,singer2022makeavideo} extended image diffusion to video by jointly modeling spatial and temporal dimensions, typically requiring end-to-end training on large video corpora. Video LDM~\cite{blattmann2023videoldm} demonstrated that temporal layers could be added to pretrained image LDMs, training only the temporal components while keeping spatial layers frozen. This paradigm was further developed by AnimateDiff~\cite{guo2023animatediff}, which proposed domain-agnostic motion modules that can be combined with various personalized text-to-image models. More recent work such as Stable Video Diffusion~\cite{blattmann2023svd} and ModelScopeT2V~\cite{wang2023modelscopet2v} has continued to advance temporal modeling within the latent diffusion framework.

\subsection{Motion-Aware Generation}

Several works have explored conditioning video generation on motion signals. MotionCtrl~\cite{wang2024motionctrl} conditions on camera trajectories, while DragAnything~\cite{wu2024draganything} allows point-based motion control. MCDiff~\cite{chen2023mcdiff} uses stroke-based motion conditioning for controllable video synthesis. However, these methods require explicit motion inputs at inference time. Our approach instead estimates motion characteristics from the content itself and adapts the temporal modeling accordingly, requiring no additional user input beyond the text prompt.

\section{Method}
\label{sec:method}

\subsection{Overview}

\begin{figure*}[t]
\begin{center}
  \includegraphics[width=0.95\textwidth]{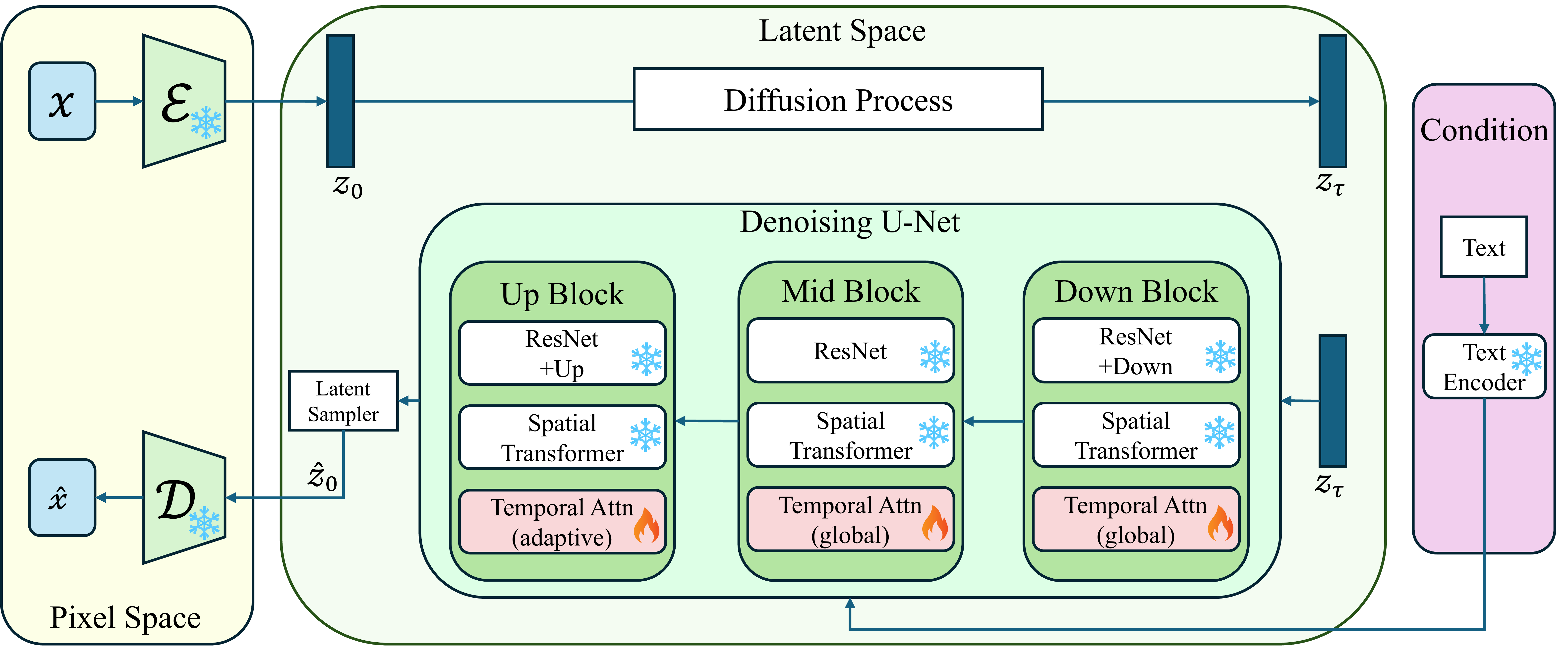}
\end{center}
\caption{Architecture overview. The overall pipeline follows the Latent Diffusion Model framework: video frames are encoded into latent space by a frozen VAE encoder~($\mathcal{E}$), processed by the denoising UNet with injected temporal attention modules (shown in red), conditioned on text via CLIP, and the VAE decoder~$\mathcal{D}$ reconstructs the frames back to pixel space. All spatial components (ResNet, Spatial Transformer) are frozen (\textit{snowflake}); only temporal attention blocks are trained (\textit{flame}). Down and mid blocks use global temporal attention; up blocks use motion-adaptive attention.}
\label{fig:architecture}
\end{figure*}

Figure~\ref{fig:architecture} illustrates the overall pipeline. Our method builds upon a pretrained Stable Diffusion model, which operates in the latent space of a frozen VAE. During training, input video frames are reshaped to $\mathbf{x} \in \mathbb{R}^{(B \cdot T) \times C \times H \times W}$ (batch size $B$, $T{=}8$ frames, $C{=}3$ RGB channels, $H{=}W{=}256$) and encoded by the VAE encoder $\mathcal{E}$ into latent representations $\mathbf{z}_0 \in \mathbb{R}^{(B \cdot T) \times c \times h \times w}$, where $h{=}w{=}32$, $c{=}4$. Noise is added via the forward diffusion process, and a UNet denoiser $\epsilon_\theta$ learns to predict the added noise. Text conditioning is provided through a frozen CLIP text encoder via cross-attention. At inference, the denoising process iteratively removes noise from random latents, and the VAE decoder $\mathcal{D}$ reconstructs video frames from the denoised latents.

Our key modification is the injection of lightweight, trainable temporal attention modules into the UNet blocks. Within each UNet transformer block, the original frozen spatial processing (spatial self-attention, text cross-attention, and feed-forward layers) is preserved as a black box. After the spatial transformer produces per-frame features, our temporal attention block operates across the frame dimension, enabling inter-frame communication at every denoising step. This is wrapped in a \texttt{SpatialTemporalWrapper} that is transparent to the rest of the UNet.

\subsection{Temporal Attention Block}
\label{sec:temporal_block}

\begin{figure}[t]
\begin{center}
  \begin{minipage}[b]{0.47\columnwidth}
    \includegraphics[width=\textwidth]{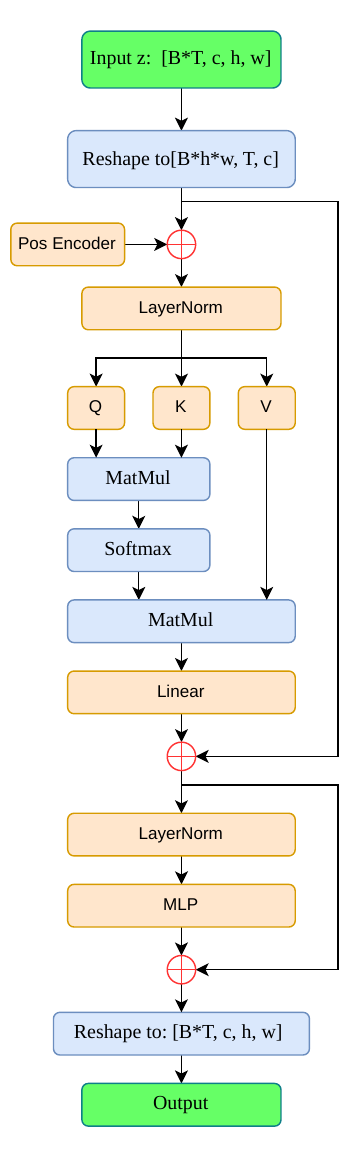}
  \end{minipage}
  \hfill
  \begin{minipage}[b]{0.47\columnwidth}
    \includegraphics[width=\textwidth]{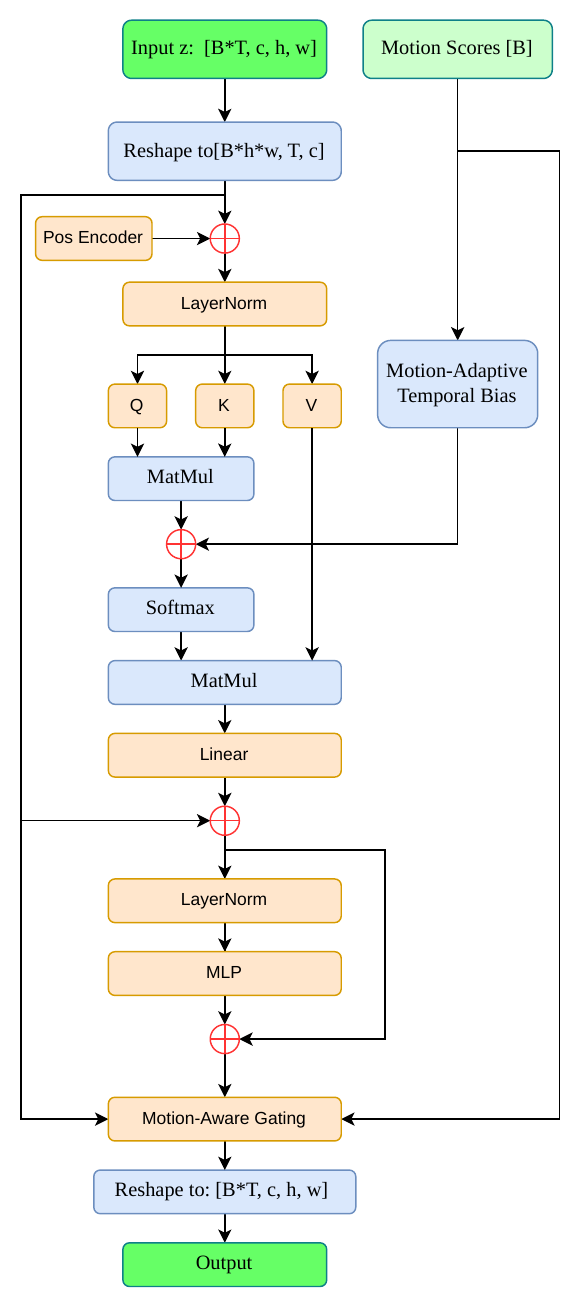}
  \end{minipage}
\end{center}
\caption{Temporal attention block variants. \textit{Left}: Global mode (down/mid blocks) --- uniform temporal self-attention, no motion conditioning. \textit{Right}: Adaptive mode (up blocks) --- motion-adaptive attention bias $\mathbf{b}_m$ and motion-aware gate $g(m)$ are both conditioned on the per-video motion score $m$.}
\label{fig:temporal_blocks}
\end{figure}

Each injected temporal attention block operates on the intermediate UNet feature tensor after spatial processing. Let $\mathbf{z} \in \mathbb{R}^{(B \cdot T) \times c \times h \times w}$ denote these features, which are reshaped to $\tilde{\mathbf{z}} \in \mathbb{R}^{(B \cdot h \cdot w) \times T \times c}$ so that attention is computed across the temporal dimension $T$ independently at each spatial position. Figure~\ref{fig:temporal_blocks} illustrates the two variants.

\subsubsection{Global Temporal Attention}

The global block applies standard multi-head temporal self-attention uniformly across all frames. Learnable temporal positional encodings $\mathbf{P} \in \mathbb{R}^{1 \times T \times c}$ are added to the input before attention, giving $\hat{\mathbf{z}} = \tilde{\mathbf{z}} + \mathbf{P}$. The block then computes:
\begin{align}
\mathbf{z}^{(1)} &= \tilde{\mathbf{z}} + \operatorname{Attn}\!\left(\operatorname{LN}(\hat{\mathbf{z}})\right) \label{eq:global_attn} \\
\mathbf{z}^{(2)} &= \mathbf{z}^{(1)} + \operatorname{FFN}\!\left(\operatorname{LN}(\mathbf{z}^{(1)})\right) \label{eq:global_ffn}
\end{align}
where $\operatorname{LN}(\cdot)$ is layer normalization, $\operatorname{Attn}(\cdot)$ is multi-head self-attention along the temporal dimension $T$, and $\operatorname{FFN}(\cdot)$ is a two-layer feed-forward network. The output $\mathbf{z}^{(2)}$ is reshaped back to $(B \cdot T) \times c \times h \times w$.

\subsubsection{Adaptive Temporal Attention}

The adaptive block extends the global formulation with two motion-conditioned components, given motion score $m \in [0,1]$ estimated from the video content.

\textbf{Motion-adaptive attention bias.} A distance-based bias matrix $\mathbf{b}_m \in \mathbb{R}^{T \times T}$ is injected into the attention logits, where $i,j \in \{1,\ldots,T\}$ index the query and key frames:
\begin{align}
\mathbf{b}_m(i,j) &= -\lambda(m) \cdot |i - j|, \qquad \lambda(m) = 0.05 + 0.45\,m \label{eq:bias}\\
\mathbf{z}^{(1)} &= \tilde{\mathbf{z}} + \operatorname{Attn}\!\left(\operatorname{LN}(\hat{\mathbf{z}}),\; \mathbf{b}_m\right) \label{eq:adaptive_attn}
\end{align}
where $\lambda(m)$ is the motion-dependent decay coefficient. When $m \to 1$ (high motion), $\lambda$ is large and attention decays steeply with frame distance, focusing each position on nearby frames. When $m \to 0$ (low motion), $\lambda$ is small and attention spans all frames globally.

\textbf{Motion-aware gating.} After the feed-forward step (identical to Eq.~\ref{eq:global_ffn}), a learned scalar gate $g(m) \in (0,1)$ modulates the strength of the temporal residual:
\begin{equation}
\mathbf{z}_{\text{out}} = \tilde{\mathbf{z}} + g(m)\cdot\left(\mathbf{z}^{(2)} - \tilde{\mathbf{z}}\right), \qquad g(m) = \sigma\!\left(\phi(m)\right)
\label{eq:gating}
\end{equation}
where $\sigma(\cdot)$ is the sigmoid function and $\phi(m)$ is a two-layer MLP with scalar output. The gate parameters are initialized such that $g(m) \approx 0$, so the block begins as a near-identity mapping and gradually learns its temporal contribution.

\subsubsection{Zero Initialization}

All output projection layers ($W_O$ and the final FFN layer) are zero-initialized. Combined with the residual connections, this ensures that at initialization both block variants are exact identity mappings, following the principle established in ControlNet~\cite{zhang2023controlnet}, and do not disturb the pretrained UNet weights at the start of training.

\subsection{Cascaded Injection Strategy}
\label{sec:cascaded}

We inject temporal attention into all transformer blocks of the UNet, but with a cascaded strategy that assigns different modes based on network depth. Down-sampling and middle blocks use \textbf{global mode}: standard temporal self-attention (no distance bias) stabilizes the semantic content and background consistency across frames at lower spatial resolutions. Up-sampling blocks use \textbf{adaptive mode}: motion-adaptive attention bias protects fine-grained features from temporal blurring, particularly in regions with significant motion. This design is motivated by the observation that low-resolution features encode scene-level semantics that should be consistent across all frames, while high-resolution features encode local details that are sensitive to inter-frame motion.

\subsection{Temporally Correlated Noise}
\label{sec:corr_noise}

Standard diffusion training and inference use independently sampled Gaussian noise for each frame. However, when the UNet processes each frame largely independently (before temporal attention takes effect), independent noise leads to completely divergent scene content across frames.

We introduce temporally correlated noise via an autoregressive process:
\begin{equation}
\label{eq:corr_noise}
\boldsymbol{\epsilon}_f = \rho \cdot \boldsymbol{\epsilon}_{f-1} + \sqrt{1 - \rho^2} \cdot \boldsymbol{\xi}_f, \quad \boldsymbol{\xi}_f \sim \mathcal{N}(\mathbf{0}, \mathbf{I})
\end{equation}
where $f \in \{1,\ldots,T\}$ is the frame index, $\rho \in [0, 1]$ is the correlation coefficient, and $\boldsymbol{\epsilon}_0 \sim \mathcal{N}(\mathbf{0}, \mathbf{I})$. Each $\boldsymbol{\epsilon}_f$ remains marginally standard Gaussian, but adjacent frames share correlated noise structure. This provides a strong prior for scene consistency without constraining the learned temporal dynamics.

Correlated noise is applied during both training and inference. The correlation coefficient $\rho$ controls a trade-off between scene consistency and motion amplitude, with $\rho = 0.6$ used during training and varied at inference for different generation behaviors (see ablation in Section~\ref{sec:ablation_noise}).

\subsection{Training Objective}
\label{sec:training_obj}

The primary training objective is the standard denoising loss:
\begin{equation}
\label{eq:denoise_loss}
\mathcal{L}_{\text{denoise}} = \mathbb{E}_{\mathbf{z}_0, \boldsymbol{\epsilon}, t} \left[ \| \boldsymbol{\epsilon} - \epsilon_\theta(\mathbf{z}_t, t, \mathbf{c}) \|^2 \right]
\end{equation}
where $\mathbf{z}_0$ are the clean latent encodings of video frames, $t$ is the diffusion timestep, $\boldsymbol{\epsilon} = [\boldsymbol{\epsilon}_1, \ldots, \boldsymbol{\epsilon}_T]$ is the temporally correlated noise from Eq.~\ref{eq:corr_noise}, and $\mathbf{z}_t = \sqrt{\bar{\alpha}_t}\,\mathbf{z}_0 + \sqrt{1-\bar{\alpha}_t}\,\boldsymbol{\epsilon}$ is the noisy latent at timestep $t$. $\mathbf{c}$ is the text conditioning. Only the temporal attention parameters are updated; the base UNet, VAE, and text encoder remain frozen.

We additionally explored a temporal consistency loss that penalizes inter-frame differences in the predicted clean latents, weighted by both motion intensity and noise level:
\begin{equation}
\label{eq:temp_loss}
\mathcal{L}_{\text{temp}} = \mathbb{E}_{t,f}\!\left[ w_m \cdot w_t \cdot \| \hat{\mathbf{z}}_0^{(f+1)} - \hat{\mathbf{z}}_0^{(f)} \|^2 \right]
\end{equation}
where $f \in \{1,\ldots,T{-}1\}$ is the frame index, $\hat{\mathbf{z}}_0^{(f)} = \left(\mathbf{z}_t^{(f)} - \sqrt{1-\bar{\alpha}_t}\,\boldsymbol{\epsilon}_f^{\,\theta}\right)/\sqrt{\bar{\alpha}_t}$ is the clean latent of frame $f$ estimated via the posterior mean, $w_m = 0.1 + 0.9 \cdot (1 - m)$ assigns higher weight to low-motion videos, and $w_t = \bar{\alpha}_t$ down-weights the loss at high noise levels. The total loss is:
\begin{equation}
\label{eq:total_loss}
\mathcal{L} = \mathcal{L}_{\text{denoise}} + \lambda_{\text{temp}} \cdot \mathcal{L}_{\text{temp}}
\end{equation}

As shown in our ablation (Section~\ref{sec:ablation_loss}), we find that the denoising loss alone ($\lambda_{\text{temp}} = 0$) produces better results than including the temporal consistency term, as the temporal attention modules learn sufficient implicit temporal regularization from the video training data.

\subsection{Motion Estimation}
\label{sec:motion_est}

During training, per-video motion scores are computed from the input frames as the mean absolute pixel difference between consecutive frames, normalized to $[0, 1]$ using dataset-level statistics (5th and 95th percentiles for robust normalization). At inference, the motion level is estimated automatically from prompt semantics---keywords such as ``running'' or ``dancing'' map to high motion, while ``still'' or ``landscape'' map to low---or can be set manually by the user.

\section{Experiments}
\label{sec:experiments}

\subsection{Setup}

\subsubsection{Architecture}

Temporal attention blocks are injected into all 16 transformer blocks of the UNet: 6 in the down-sampling path, 1 in the middle block, and 9 in the up-sampling path. Each temporal block uses 4 attention heads with dimension 64 per head and a feed-forward inner dimension of 512. This adds 25.8M trainable parameters (2.9\% of the 860M base UNet). We evaluate with two frozen base models: SD~1.5 (\texttt{sd-legacy/stable-diffusion-v1-5}) and SD~2.1-base (\texttt{stabilityai/stable-diffusion-2-1-base}), the 512${\times}$512 variant distinct from the 768${\times}$768 \texttt{stable-diffusion-2-1} model. Our main comparison uses SD~2.1-base.

\subsubsection{Training Data}

We train on a 100K-video subset of WebVid-10M~\cite{bain2021webvid} for all experiments. We sample 8 frames per video at a frame interval of 2 and resize to $256{\times}256$. Motion scores are pre-computed for the entire training set using robust percentile normalization.

\subsubsection{Training Details}

Training uses AdamW with learning rate $1{\times}10^{-4}$, cosine schedule with 500 warm-up steps, and gradient norm clipping at~1.0. We train for 30 epochs with batch size~55 per GPU across A100 GPUs using mixed-precision (FP16) and gradient checkpointing. The noise correlation coefficient is $\rho = 0.6$ during training.

\subsubsection{Inference}

We use DPM-Solver++~\cite{lu2022dpm} with Karras noise schedule for 50 denoising steps, classifier-free guidance scale of~7.5, and resolution $256{\times}256$ with 8 frames per video. The noise correlation coefficient is adjustable at inference time without retraining.

\subsection{Evaluation Metrics}

We evaluate on 400 uniformly sampled captions from the WebVid validation set using the following metrics. \textbf{CLIP-SIM}: average CLIP~\cite{radford2021clip} cosine similarity between each frame and its text prompt, measuring text-video alignment. \textbf{Frame Consistency (FC)}: average CLIP cosine similarity 
between consecutive frame embeddings, measuring temporal coherence; 
a very low FC indicates incoherence, while an excessively high FC
may reflect static output rather than good generation quality.
\textbf{Motion Magnitude (MM)}: mean absolute pixel difference between consecutive frames in pixel space, measuring motion amplitude. \textbf{FVD}: Fr\'{e}chet Video Distance~\cite{unterthiner2019fvd} measuring overall video quality against real WebVid videos. \textbf{FID}: Fr\'{e}chet Inception Distance~\cite{heusel2017fid} computed per-frame, measuring single-frame image quality.

\subsection{Comparison with AnimateDiff}
\label{sec:main_results}

Table~\ref{tab:main_results} compares our method against AnimateDiff~\cite{guo2023animatediff} under equal training conditions. To ensure a fair comparison, we retrain AnimateDiff's motion modules on the same 100K-video WebVid subset with identical settings ($T{=}8$ frames, $256{\times}256$, same evaluation protocol). We also include a baseline using SD~1.5 with correlated noise but without temporal modules to isolate the contribution of our temporal attention. Removing the temporal modules causes FID to rise from 89.96 to 180.48 and FVD from 85.99 to 211.02, while frame consistency drops sharply from 0.980 to 0.777, confirming that temporal attention is the primary driver of inter-frame coherence. AnimateDiff achieves better FID and FVD under these equal-data conditions, which we attribute to its larger motion module capacity (417M parameters). Importantly, our temporal modules add only 25.8M trainable parameters---$16{\times}$ fewer---making our approach substantially more parameter-efficient.

\begin{table}[t]
\caption{Comparison with AnimateDiff on WebVid validation (400 samples, $\rho{=}0.6$, $\lambda_{\text{temp}}{=}0$, 100K training videos). $\dagger$~AnimateDiff retrained on the same 100K WebVid subset with $T{=}8$, $256{\times}256$, matching inference settings. $\ddagger$~Base SD~1.5 without temporal modules; high MM reflects incoherent frame-independent generation.}
\label{tab:main_results}
\begin{center}
\begin{tabular}{l c c c c c}
\toprule
Method & CLIP$\uparrow$ & FC & MM$\uparrow$ & FID$\downarrow$ & FVD$\downarrow$ \\
\midrule
AnimateDiff$\dagger$ & \textbf{0.308} & 0.969 & \textbf{0.062} & \textbf{81.63} & \textbf{79.61} \\
SD~1.5$\ddagger$ & 0.267 & 0.777 & 0.326 & 180.48 & 211.02 \\
Ours (SD~1.5) & 0.302 & 0.980 & 0.051 & 89.96 & 85.99 \\
Ours (SD~2.1) & 0.303 & 0.983 & 0.040 & 92.09 & 92.49 \\
\bottomrule
\end{tabular}
\end{center}
\end{table}

\subsection{Ablation Studies}

Unless otherwise noted, ablation experiments use SD~2.1 trained on the 100K subset, evaluated on 400 WebVid validation captions. The base model comparison (Table~\ref{tab:ablation_sd}) additionally includes SD~1.5 under identical settings.

\subsubsection{Effect of Temporal Consistency Loss}
\label{sec:ablation_loss}

Table~\ref{tab:ablation_loss} compares training with the combined loss ($\lambda_{\text{temp}} = 2.0$) versus denoising loss only ($\lambda_{\text{temp}} = 0$). Removing the explicit temporal loss produces videos with larger motion amplitude ($0.040$ vs.\ $0.025$). The temporal attention mechanism provides sufficient implicit regularization through the denoising objective applied to temporally coherent video data, making the explicit consistency term unnecessary and even counterproductive for motion diversity.

\begin{table}[t]
\caption{Effect of temporal consistency loss ($\lambda_{\text{temp}}$). SD~2.1, 100K, $\rho{=}0.6$.}
\label{tab:ablation_loss}
\begin{center}
\begin{tabular}{l c c c c c}
\toprule
$\lambda_{\text{temp}}$ & CLIP$\uparrow$ & FC & MM$\uparrow$ & FID$\downarrow$ & FVD$\downarrow$ \\
\midrule
2.0 & \textbf{0.305} & 0.990 & 0.025 & 95.82 & 97.21 \\
\textbf{0} & 0.303 & 0.983 & \textbf{0.040} & \textbf{92.09} & \textbf{92.49} \\
\bottomrule
\end{tabular}
\end{center}
\end{table}

\subsubsection{Effect of Noise Correlation}
\label{sec:ablation_noise}

Table~\ref{tab:ablation_noise} reports the effect of the noise correlation coefficient $\rho$ at inference time, using a model trained with $\rho{=}0.6$. Higher correlation suppresses motion amplitude: at $\rho{=}0.85$, motion magnitude collapses to $0.008$ (effectively static video). Lower correlation allows more diverse inter-frame content. This parameter provides a user-controllable trade-off at inference time without retraining.

\begin{table}[t]
\caption{Effect of noise correlation $\rho$ at inference. SD~2.1, 100K, $\lambda_{\text{temp}}{=}0$.}
\label{tab:ablation_noise}
\begin{center}
\begin{tabular}{c c c c c c}
\toprule
$\rho$ & CLIP$\uparrow$ & FC & MM$\uparrow$ & FID$\downarrow$ & FVD$\downarrow$ \\
\midrule
0.4 & 0.278 & 0.929 & \textbf{0.158} & 105.85 & 108.22 \\
\textbf{0.6} & \textbf{0.303} & 0.983 & 0.040 & \textbf{92.09} & \textbf{92.49} \\
0.85 & 0.298 & 0.999 & 0.008 & 130.87 & 140.11 \\
\bottomrule
\end{tabular}
\end{center}
\end{table}

\subsubsection{Effect of Base Model}
\label{sec:ablation_sd}

Table~\ref{tab:ablation_sd} compares Stable Diffusion~1.5 and~2.1 as base models under otherwise identical settings (100K training, $\rho{=}0.6$, $\lambda_{\text{temp}}{=}0$). SD~1.5 achieves slightly better FID and FVD, likely because the WebVid reference statistics are closer to SD~1.5's training distribution, and produces slightly larger motion amplitude. Both models benefit substantially from temporal attention modules.

\begin{table}[t]
\caption{Effect of base model. 100K, $\rho{=}0.6$, $\lambda_{\text{temp}}{=}0$.}
\label{tab:ablation_sd}
\begin{center}
\begin{tabular}{l c c c c c}
\toprule
Base Model & CLIP$\uparrow$ & FC & MM$\uparrow$ & FID$\downarrow$ & FVD$\downarrow$ \\
\midrule
SD~1.5 & 0.302 & 0.980 & \textbf{0.051} & \textbf{89.96} & \textbf{85.99} \\
SD~2.1 & \textbf{0.303} & 0.983 & 0.040 & 92.09 & 92.49 \\
\bottomrule
\end{tabular}
\end{center}
\end{table}

\subsection{Qualitative Results}

\begin{figure}[t]
\centering
\setlength{\tabcolsep}{1pt}
\renewcommand{\arraystretch}{0.5}
\small
\begin{tabular}{r l}

\multicolumn{2}{l}{\small\textit{(a) ``Time lapse of sky clouds storm''}} \\[2pt]
\raisebox{0.4\height}{\footnotesize Ours (SD\,1.5)\;} &
  \includegraphics[height=3.0cm]{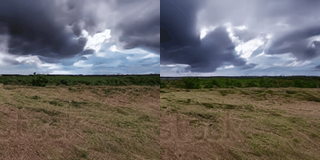} \\[2pt]
\raisebox{0.4\height}{\footnotesize Ours (SD\,2.1)\;} &
  \includegraphics[height=3.0cm]{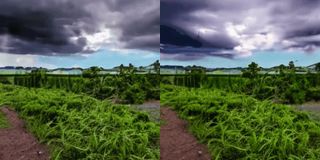} \\[2pt]
\raisebox{0.4\height}{\footnotesize AnimateDiff$\dagger$\;} &
  \includegraphics[height=3.0cm]{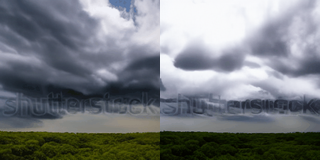} \\[3pt]

\multicolumn{2}{l}{\small\textit{(b) ``Pacific black duck, anas superciliosa, relaxing''}} \\[2pt]
\raisebox{0.4\height}{\footnotesize Ours (SD\,1.5)\;} &
  \includegraphics[height=3.0cm]{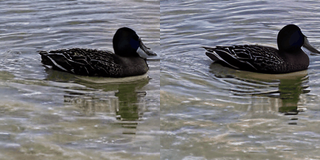} \\[2pt]
\raisebox{0.4\height}{\footnotesize Ours (SD\,2.1)\;} &
  \includegraphics[height=3.0cm]{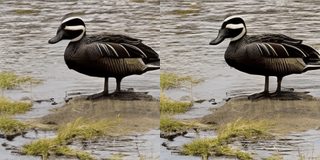} \\[2pt]
\raisebox{0.4\height}{\footnotesize AnimateDiff$\dagger$\;} &
  \includegraphics[height=3.0cm]{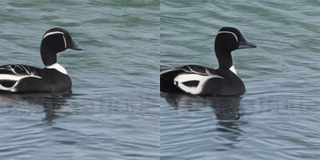} \\

\end{tabular}
\caption{Qualitative comparison on two WebVid validation prompts. Each pair shows frame~1 and frame~8 of the 8-frame generated sequence at $256{\times}256$, highlighting temporal consistency across the full generation. AnimateDiff$\dagger$  retrained on 100K videos, $T{=}8$.}
\label{fig:qualitative}
\end{figure}

Figure~\ref{fig:qualitative} shows the first and last frame (frames 1 and 8) of generated sequences from our method and the retrained AnimateDiff baseline for two representative WebVid validation prompts. Displaying the temporal endpoints highlights scene consistency: a method with poor temporal coherence would show visible subject or background drift between frame 1 and frame 8. Our method maintains consistent subject identity and scene appearance across the full sequence, with natural motion appropriate to the content.

\section{Discussion}
\label{sec:discussion}

\textbf{Implicit vs.\ explicit temporal regularization.}
A key finding is that the denoising loss alone suffices for temporal learning, and explicit temporal smoothness losses are counterproductive. We hypothesize that the temporal attention modules learn to produce consistent noise predictions because inconsistent predictions are penalized by the denoising loss when trained on video frames that are inherently temporally coherent. The explicit temporal loss over-constrains the system by penalizing all inter-frame variation, suppressing natural motion diversity.

\textbf{Role of correlated noise.}
Temporally correlated noise serves as a crucial initialization prior. Without it, the UNet---which processes each frame through frozen spatial attention independently---produces completely divergent scene content across frames, creating an impossible task for the lightweight temporal attention to correct. Correlated noise ensures that frames start from similar noise patterns, providing a foundation that the temporal attention can then refine. The correlation coefficient provides a practical inference-time knob that does not require retraining.

\textbf{Motion amplitude trade-off.}
Our ablations reveal that $\rho$ directly controls motion amplitude. At $\rho{=}0.85$, videos are effectively static (MM$\,{=}\,0.008$); at $\rho{=}0.4$, motion is rich (MM$\,{=}\,0.158$) but at some cost to CLIP alignment (0.278). The operating point $\rho{=}0.6$ strikes a practical balance for diverse video prompts.

\textbf{Limitations.}
Our method generates 8 frames at $256{\times}256$, which is below the resolution and duration of state-of-the-art video generators trained on orders of magnitude more data. The FID and FVD scores remain above AnimateDiff's even under equal training data, reflecting the capacity advantage of AnimateDiff's larger motion modules. Future work will explore scaling to higher resolutions, longer sequences, and integration with more modern architectures such as DiT-based diffusion models~\cite{peebles2023dit}.

\section{Conclusion}
\label{sec:conclusion}

We presented a motion-adaptive temporal attention mechanism for parameter-efficient video generation with frozen Stable Diffusion models. By injecting lightweight temporal modules (25.8M parameters, 2.9\% of the base UNet) with motion-adaptive attention bias, motion-aware gating, and a cascaded global-to-adaptive injection strategy, our method produces temporally coherent videos while adapting to diverse motion characteristics. Combined with temporally correlated noise, the approach achieves competitive results with $16{\times}$ fewer parameters than AnimateDiff under identical training conditions (100K videos, $T{=}8$). Our analysis reveals that implicit temporal regularization through the denoising objective alone outperforms explicit temporal consistency losses, and that the noise correlation coefficient provides a convenient inference-time control for balancing consistency and motion amplitude---a trade-off not typically exposed by fixed-architecture methods.



\begin{biography}
\textbf{Rui Hong} is a PhD student at George Mason University. His research interests include sign language translation and generation, 3D pose estimation, multimodality AI, and generative AI.

\textbf{Shuxue Quan} is an independent researcher with interests in 
artificial intelligence, computer vision, camera systems, and AR/VR/XR/MR technologies.
\end{biography}

\end{document}